\documentclass[pre,
showpacs,floatfix,
twocolumn
]{revtex4-1} 

\usepackage{amsfonts}
\usepackage{amsmath,amssymb}
\usepackage{cancel,verbatim}
\usepackage{graphicx,float}
\usepackage{color}
\usepackage{bm,subfigure}
\usepackage{epsfig}
\usepackage{todonotes}
\usepackage{hyperref} 

\newcommand{\beq}{\begin{equation}}  
\newcommand{\eeq}{\end{equation}}  
\newcommand{\bea}{\begin{eqnarray}}  
\newcommand{\eea}{\end{eqnarray}}  

\renewcommand{\vec}[1]{\bm{#1}}

\newcommand{\review}[1]{#1}

\begin{document}

\title{Using Machine Learning to predict extreme events in the H\'{e}non map}

\author{Martin Lellep\textsuperscript{1}}
\email{martin.lellep@physik.uni-marburg.de}
\author{Jonathan Prexl\textsuperscript{2}, Moritz Linkmann\textsuperscript{1}}
\author{Bruno Eckhardt\textsuperscript{1}}
\email{Deceased on the $7^{\rm th}$ of August 2019.}
\affiliation{
	\textsuperscript{1}Physics Department, Philipps-University of Marburg, D-35032 Marburg, Germany \\
	\textsuperscript{2}Department of Civil, Geo and Environmental Engineering, Technical University of Munich, D-80333 Munich, Germany
}

\date{\today}

\begin{abstract}
Machine Learning (ML) inspired algorithms provide a flexible set of tools for analyzing
and forecasting chaotic dynamical systems. We here analyze the performance of
	one algorithm for the prediction of extreme events in the 
	two-dimensional
	H\'{e}non map at the classical
parameters. The task is to determine whether a trajectory will exceed a threshold 
after a set number of 
time steps into the future. This task has a geometric interpretation within the dynamics
of the H\'{e}non map, which we use to gauge the performance of the neural networks that 
are used in this work. We analyze the dependence of the success rate of the ML models 
on the prediction time $T$, the number of training samples $N_T$ and the size of the network $N_p$. 
We \review{observe} that in order to maintain a certain accuracy, $N_T\propto \exp(2hT)$ and 
$N_p \propto \exp(hT)$, where $h$ is the topological entropy. 
Similar relations between the intrinsic chaotic properties of the dynamics and ML parameters
\review{might be observable} in other systems as well.
\end{abstract}


\maketitle

\textbf{The power of ML algorithms in the handling of complex tasks is impressively
demonstrated by their successes in the game of Go \cite{Silver:2017bo} and real world applications
from pattern recognition to autonomous driving. Within physics, they are used to 
select suitable bases for many body quantum systems \cite{viewpoint}, to classify patterns and phases in 
many body systems \cite{jeckel2019learning} or phase transitions in material 
science \cite{carrasquilla2017machine,butler2018machine}, 
and to derive low-dimensional models for turbulent flows \cite{brunton2019machine}
and other systems \cite{Champion:2019vq}, to mention just a few. 
In chaotic systems, time series prediction is challenging because
of the sensitive dependence on initial conditions and the ensuing increase in uncertainty
\cite{ott_2002}. We here analyze the performance of a particular class of ML algorithms for
predicting extreme events in the H\'{e}non map, as a model for similar tasks in other
systems.}

\section{Introduction}
\label{sec:intro}

Machine Learning (ML) algorithms emerge as a flexible and versatile tool for the analysis and the 
characterization of complex systems. Following the initial applications of neural networks
for the prediction of a wide range of physical and non-physical time series
\cite{weigend,svm,Bontempi2013}, interest 
in ML has been revived by the impressive demonstration of the 
representation and prediction of a spatially-extended Kuramoto-Shivashinsky system by 
Ott and collaborators \cite{pathak2018model}. 
They have also shown, how the dynamics of a trained network can be used to extract the Lyapunov
exponents for the original system from the Lyapunov exponent of the network \cite{Pathak:2017io}.
As prediction is a common task in many situations, including targeting
satellites, forecasting weather \cite{lorenz1963deterministic} or stock markets, ML learning tools hold great
potential because of their apparently high adaptability despite rather simple design principles.
However, there are numerous parameters that influence the power of the networks,
including the number of training samples, the time interval over which a prediction is requested,
or the topology of the network. Our aim here is to analyze the effect of some of these
parameters.

The task we use to gauge the power of ML algorithms is guided by considerations in transitional
shear flows: experimental, numerical and theoretical results show that in parameter
ranges intermediate between laminar and fully developed turbulent, the turbulent
states are not persistent but transient \cite{Hof:2006ab}. The decay is memoryless, and happens without
any noticeable precursor. Nevertheless, decaying trajectories have to depart from the
turbulent \review{saddle} at some point, and the identification of when and where this happens
can be useful for manipulations of the turbulent dynamics and the understanding of the 
turbulent \review{saddle} itself (see, e.g. \cite{Eckhardt:2007ka,Eckhardt:2018js}). 
Here we rephrase the decay prediction 
problem as a task in the H\'{e}non map, and leave the application to the relaminarization
problem (see Schlatter et al \cite{Schlatter:2019} for an example) for a future publication.

We use the 
two-dimensional (2d)
H\'{e}non map as dynamical system and pose a task that has a geometric 
interpretation in the state space of the system: the task is to predict whether a trajectory that 
starts at a certain point in state space will pass a prescribed threshold $\theta$ after $T$ iterations. 
In relation to the relaminarization problem in transient turbulent flows, the threshold corresponds to
a barrier beyond which the flow becomes laminar. When reduced
to two dimensions and a map, the task becomes much easier to analyze, while at the 
same time keeping the essential dynamical features of the flow: the Lyapunov 
exponent is positive, the task becomes more challenging with increasing time, 
and it depends on ever finer details of the initial conditions. 

The outline of the paper is as follows. 
In section 2 we describe the H\'{e}non map and the geometrical interpretation of the task.
In section 3, we discuss the training of the network and its dependence on 
three critical parameters: the number of training samples $N_T$, the forecasting time $T$,
and the number of parameters $N_p$ in the neural network used for the task.
We conclude with a few observations and general remarks in section 5 after presenting the results in section 4.

\section{Thresholds in the H\'{e}non map}
\label{sec:model}

\begin{figure}
	\centering
	\includegraphics[scale=1]{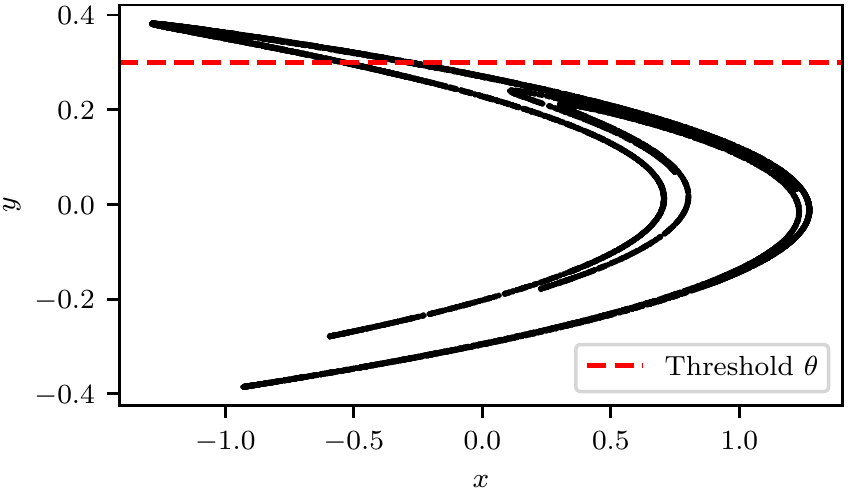}
	\caption{The H\'{e}non attractor in state space. Points on the  H\'{e}non attractor are shown as black dots and
	the dashed red line denotes the threshold $\theta$ that is used 
	for the prediction task in Eq.~\eqref{eq:criterion}.
    }
	\label{fig:phase_space}
\end{figure}

The H\'{e}non map \cite{henon1976} is a two dimensional discrete non-linear map, 
\begin{align}
\begin{split}
	x_{n+1} & = 1 - a x_n^2 + y_n \\
	y_{n+1} & = b x_n.
	\label{eq:henon}
\end{split}
\end{align}
with two parameters $a$ and $b$. For the common choice of $a=1.4$ and $b=0.3$,
the dynamics is chaotic with a positive Lyapunov exponent, and a positive topological
entropy $h=0.465$ \review{\cite{artuso1990recycling}}.
\review{Equation~\eqref{eq:henon} can be written as a one dimensional
discrete non-linear map that depends on the two preceding iterates,
$z_{n+1} = 1 - a z_{n}^2 + b z_{n-1}$. However, the two dimensional
H\'{e}non map formulation is used here.} 
The familiar shape of the attractor with the threshold is shown 
in Fig.~\ref{fig:phase_space}. 
The threshold \review{for the $y$ coordinate} is set at $\theta=0.3$ and is kept fixed for all studies
presented here: while other values will give changes in detail, we do not expect any substantial
changes in the general conclusions we draw here.

The prediction task on the chaotic attractor is defined as follows: Given as input a set of phase space points 
$(x_i, y_i)$  where $i=n$, $n-1$, $n-2$, \ldots, $n-(N-1)$, i.e. the phase space point at time $n$ and its 
predecessors down to $n-(N-1)$ for a total of $N$ points, 
\review{
\begin{equation}
	\{(x_i, y_i)\}_{i=n-(N-1), \dots, n},
	\label{eq:history}
\end{equation}
}
predict whether the macroscopic criterion
\begin{equation}
	y_{n+T}\ge\theta=0.3
	\label{eq:criterion}
\end{equation}
is fulfilled at a point in time that is $T$ steps in the future. Note that we do not ask whether 
the threshold is passed at intermediate times: while this may be relevant for the decay problem of turbulence, 
numerical results show that the main effect of this change is to fragment the state space even more.
The particular choice of $\theta$ is arbitrary, and a range of values will give similar results. For finite 
$T$, the points that pass the threshold vary continuously with $\theta$ so that the results will be
robust under variations with $\theta$. For increasing times the range in $\theta$ for which this
remains true decreases, and eventually discontinuous changes in the regions in which the criterion
is satisfied will appear.

\begin{figure}
	\centering
	\includegraphics[scale=1]{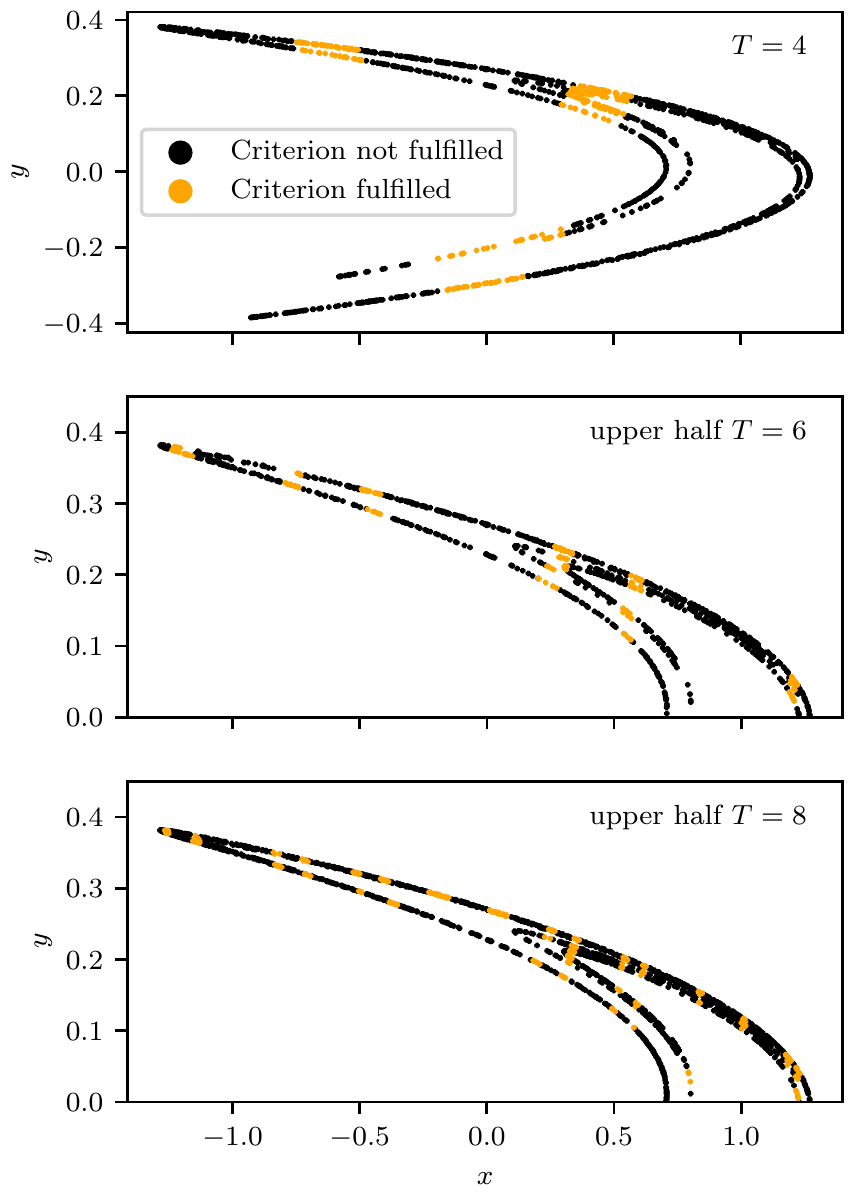}
	\caption{Points on the attractor that pass the threshold after $T$ steps for $T=4$ (top), $T=6$ (middle) and
	$T=8$ (bottom). Points on the attractor that are above the threshold are shown in orange, those below in black.
	To better demonstrate the small regions for larger $T$, the middle and bottom frame only show the 
	upper half $y\ge 0$ of the state space.}
	\label{fig:geometrical_task}
\end{figure}

Equation~\eqref{eq:criterion} resembles a macroscopic criterion in a chaotic trajectory and is motivated by the criterion we usually use in the
determination of turbulent lifetimes: trajectories decay when the energy content in transverse modes falls
below a critical value \cite{Schneider:2009cq}, here replaced by a trajectory reaching $y\ge \theta$. The regions that satisfy 
the criterion are shown in Fig.~\ref{fig:geometrical_task}:  The initial conditions that will pass the threshold
after $T$ time steps come from compact regions in the phase space that increase in number and
become smaller in diameter as $T$ increases.

\begin{figure}
	\centering
	\includegraphics[scale=1]{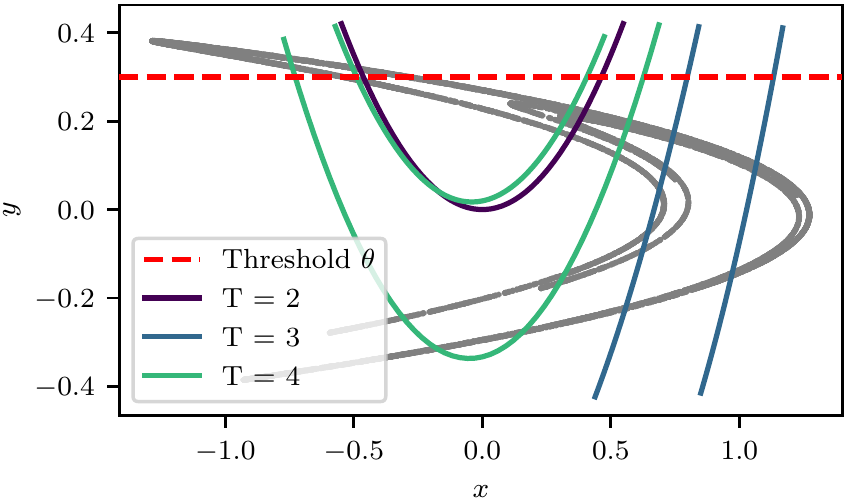}
	\caption{Inverse iterates of the threshold. The gray points show the H\'{e}non attractor and the red line marks the threshold of the criterion \eqref{eq:criterion}. Preimages of the threshold become more complex as the number of time steps $T$ into the past increases, as initially connected phase space regions are split up into subsets. The preimages of the threshold develop ever more folds that cut the H\'{e}non attractor. 
    }
	\label{fig:inverse_criterion}
\end{figure}

The regions that satisfy the criterion  \eqref{eq:criterion} after $T$ iterations have a geometrical 
interpretation, obtained by considering the preimages of the threshold, i.e. by 
asking which points end up on the threshold after $n$ steps. 
To start, the preimages of a point $(x_{n+1}, y_{n+1})$ are determined by
the inverse of the map,
\begin{align}
\begin{split}
	x_{n} & = \frac{y_{n+1}}{b} \\
	y_{n} & = x_{n+1} -1 + a \frac{y_{n+1}^2}{b^2}.
	\label{eq:inverse_henon}
\end{split}
\end{align}
Specifically, the line $y_{n+1}=\theta$ has as a preimage the line
\begin{align}
\begin{split}
	x_{n} & = \frac{\theta}{b} \\
	y_{n} & = x_{n+1} -1 + a \frac{\theta^2}{b^2}.
	\label{eq:inverse_henon_1}
\end{split}
\end{align}
Everything to the right of $x_{n}=\theta/b$ will be mapped above $y_{n+1}=\theta$. Going one
more step into the past, the line $(x_{n}=\theta/b, y_n)$ will be mapped to the parabola
\begin{align}
\begin{split}
	x_{n-1} & = \frac{y_{n}}{b} \\
	y_{n-1} & = \frac{\theta}{b} -1 + a \frac{y_{n}^2}{b^2}.
	\label{eq:inverse_henon_2}
\end{split}
\end{align}
All points inside the parabola will be mapped above $y_{n+1}=\theta$ in two steps.
Iterating further into the past, the pre-images become more convoluted, as indicated in 
Fig.~\ref{fig:inverse_criterion}. As the number of time steps into the past increases,
the preimages of the threshold develop more and more lines that cross the attractor,
\review{each one of them dividing the attractor into ever smaller segments that will be mapped
above $y_n$ for some time steps into the future.}



\section{Machine learning for predictions}
\label{sec:predict}

We now turn to the application of ML tools for the prediction task. The prediction task formulated in the last preceeding section is considered a \textit{classification task} in the ML literature \cite{goodfellow2016deep}. The neural network that is used as ML model in this work is trained to classify a given sample to belong to one of the two classes separated by the criterion Eq.~\eqref{eq:criterion}. Consequently, the regions which make up the geometrical task interpretation in Fig.~\ref{fig:geometrical_task} are considered \textit{classes} in the ML parlance. One class are the phase space points colored in orange in Fig.~\ref{fig:geometrical_task} that fulfill Eq.~\eqref{eq:criterion} and the other class are the phase space points colored in black, which do not fulfill \eqref{eq:criterion}.

Among the many possible network topologies 
and neural dynamics, we use a fully connected feed forward neural network. 

\begin{figure}
	\centering
	\includegraphics[width=\columnwidth]{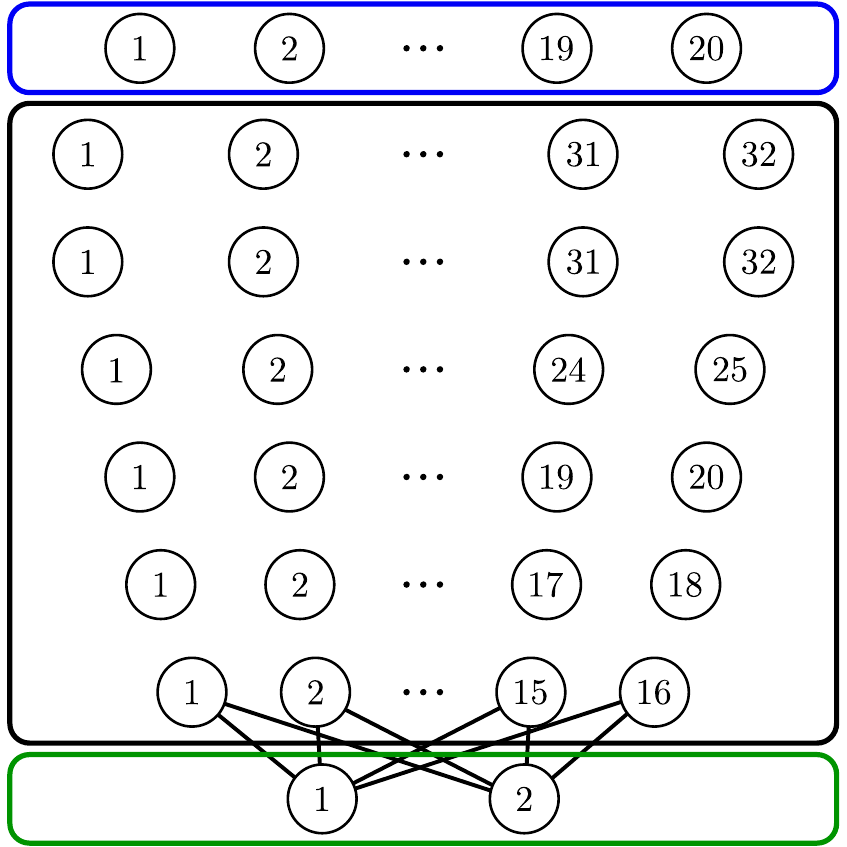}
	\caption{Topology of the neural network. There are $2N$ input nodes (blue box), with $N$ the number
	of time steps in the past, $(32, 32, 25, 20, 18, 16)$ hidden layers (black box), with a ReLU activation function,
	and 2 output neurons (green box) with a softmax activation function. The network is fully connected between layers. Each neuron of the input and hidden layers is connected to every neuron in its subsequent layer. Only the weights of the last hidden layer to the output layer are visualized in order to avoid excess visual load by the large number of weights. The numbers enumerate the neurons in each layer.}
	\label{fig:neural_network}
\end{figure}

The latter is \review{ 
a directed acyclic graph that is made up of layers, as shown
schematically in Fig.~\ref{fig:neural_network}.} \review{The
interactions between the layers $i-1$ and $i$ 
are computed 
by applying an activation function $g$
to the
latent variable $\vec{z}^{(i)}$, which itself is a linear function of
the previous layer $\vec{y}^{(i-1)}$,
\begin{align}
	\begin{split}
		\vec{z}^{(i)} & = \vec{W}^{(i, i-1)} \cdot \vec{y}^{(i-1)} \\
		\vec{y}^{(i)} & = g(\vec{z}^{(i)}).
	\end{split}
	\label{eq:NN_dynamics}
\end{align}
The weight matrix $\vec{W}^{(i, i-1)}$ connects the neurons of layer $i-1$ stored in $\vec{y}^{(i-1)}$ to the latent variables of layer $i$ stored in $\vec{z}^{(i)}$. An evaluation of a fully connected feed forward neural network then applies Eq.~\eqref{eq:NN_dynamics} for each layer in the network, starting from the input layer $i=0$ and ending with the output layer.}
\review{Feed forward neural networks are trained by adapting all weight
matrices summarized into a high dimensional parameter vector
$\mathbf{\Gamma}=[\vec{W}^{(i, i-1)}]_{i=1, \dots, n_l}$, where $n_l$ is the number of layers, 
in order to minimize a pre-defined loss functional $\mathcal{L}(\mathbf{\Gamma})$.} 

\begin{figure}
	\centering
	\includegraphics[scale=1]{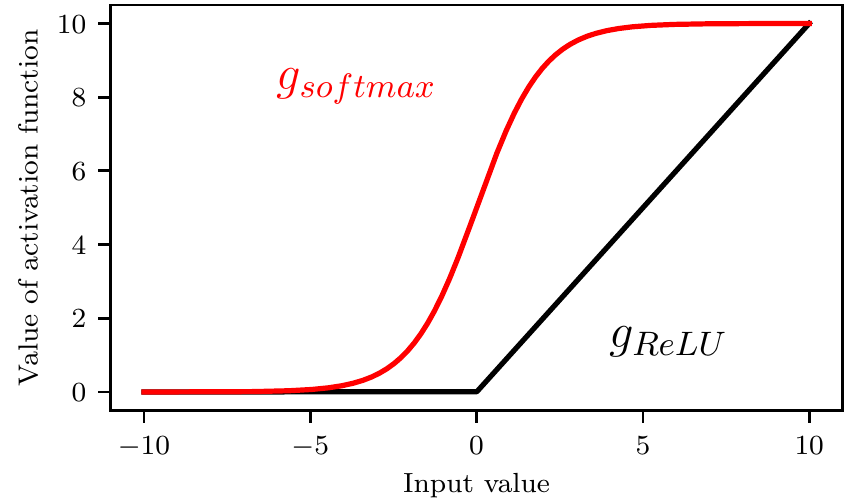}
	\caption{
	Activation functions for hidden and output layers. The ReLU function
	$ReLU(x)=\max(0, x)$ (black) is used for the hidden layers, and the softmax function (red) 
	for the output layer. Since there are two output classes, the softmax function is
        equivalent to the sigmoid function, $\sigma(x)=1/(1-e^{-x})$. The
        softmax function is scaled to the maximal value of $10$ for
        visual purposes.		
	}
	\label{fig:activation_functions}
\end{figure}

\review{The neural network we use here has $n_l=8$ layers, the first layer being the input layer,
followed by six hidden layers and a final output layer. ReLU activation functions \cite{goodfellow2016deep} are used for the hidden layers and the output layer uses a softmax activation function \cite{goodfellow2016deep}. These two types of activation functions are shown in Fig.~\ref{fig:activation_functions}.} While the details of the results we present will depend on this specific choice, we anticipate that other network topologies or activation functions will show similar behaviour.

With an input based on the current position and $N-1$ further steps into the past, it has dimensions
$2N$. Therefore, the number of neurons in the first layer is $2N$, where we work with $N=10$.
The numbers of neurons per layer are then taken to be $(2 N, 32, 32, 25, 20, 18, 16, 2)$, with the number of 
hidden nodes and the final output node fixed. \review{The output as softmax layer returns two numbers that are often interpreted as probabilities of the input being in the respective classes. The input is then classified according to which of the two numbers is larger.} The neural network is shown schematically in Fig.~\ref{fig:neural_network}. \review{Choosing the specific neural network topology being a heuristic, the first hidden layer is enlarged and subsequent layers follow roughly a linear interpolation down to the output layer, as suggested in \cite{heaton2008introduction}. The topology presented above starts to learn the task by decreasing the loss function value upon training and, hence, has been chosen as baseline topology.}

Concerning the loss functional, the
\review{
binary classification problem motivates the usage of the binary categorial
cross-entropy \cite{goodfellow2016deep} as it is commonly defined in Machine
Learning \cite{goodfellow2016deep} by
\begin{equation}
	\mathcal{L}(\mathbf{\Gamma}) = \frac{1}{N_T} \sum_{i=1}^{N_T} y_i \log(p(x_i, \mathbf{\Gamma})) + (1-y_i) \log(1-p(x_i, \mathbf{\Gamma}))
\end{equation}
with $N_T$ as the number of training samples, $y_i$ the true label of training
sample $i$, $x_i$ the trajectory data of sample $i$ and $p(x_i,
\mathbf{\Gamma})$ the probability of sample $i$ to fulfill criterion
\eqref{eq:criterion} as predicted by the neural network with parameters
$\mathbf{\Gamma}$. The Adam optimizer \cite {kingma2014adam} is used to
minimize $\mathcal{L}$. This optimizer extends the regular stochastic gradient
descent optimizer by step size adaptivity per weight and uses past gradient
magnitudes of $\mathcal{L}$ to estimate appropriate step sizes. 
Calculating the gradients of all $N_T$ summands is computationally unfeasible. 
The underlying stochastic gradient descent therefore approximates
$\nabla_{\Gamma}\mathcal{L}(\vec{\Gamma})$ by taking $m \ll N_T$ random
samples from the $N_T$ training samples.
The quantity $m$ is called batchsize and is here set to $m=128$. The
training is performed for $5000$ epochs, where one epoch consists of showing
all $N_T$ training samples once to the optimizer during the stochastic gradient
step.}

\begin{figure}
	\centering
	\includegraphics[scale=1]{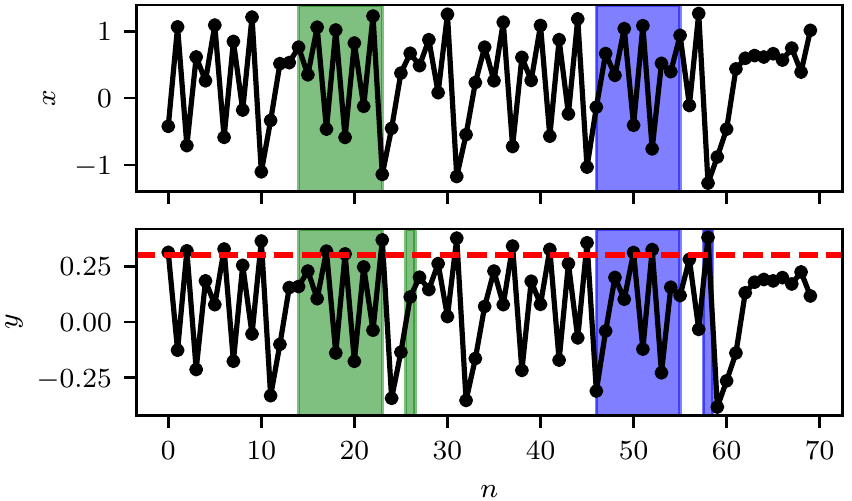}
	\caption{
	Example trajectory of the H\'{e}non map with explanation of training data construction. The two subplots correspond to the components $x$ and $y$. The black lines with dots are the iterates starting from some initial conditions. The dashed red line marks the threshold of $\theta=0.3$ for the $y$ component that is used for the task definition. Furthermore, two sets of points are colored in green and blue to exemplify how training data is constructed for samples that do not fulfill and fulfill the task criterion, respectively. The large sets of colored points correspond to the extracted training samples which are used during the training of the neural network. The small stripes only present in the $y$ subplot emphasize that the training samples are constructed based on the $y$ coordinate at future time points. The time horizon for this visualization is set to $T=3$, so that the small stripes are located $3$ iterations after the actual samples.}
	\label{fig:dynamics}
\end{figure}

The number of samples $N_T$ in the training set is varied from $10^3$ to $3\cdot 10^5$,
as detailed below. The training data is sampled from the natural H\'{e}non dynamics by first
iterating Eq.~\eqref{eq:henon} and second including each point's history according to Eq.~\eqref{eq:history},
see Fig.~\eqref{fig:dynamics}. Finally, the training data are scaled to zero-mean and unit variance.
The training labels are determined according to Eq.~\eqref{eq:criterion} and \review{$N_T/2$ training samples per each of the two classes are used in order to train the network on an equal number of samples per class. This avoids biases in the predictions of the neural network that occur for imbalanced data \cite{he2009learning}.}

The performance of the network is evaluated by the predictions for a fixed large data set 
of $10^{6}$ samples that have not been used for training purposes.
\review{Also for this test data, 
we ensure
that the number of samples per class is equal for both classes.}
The percentage of correctly classified samples of this verification
corpus is called \textit{accuracy} and is used as the measure for the performance of the network. 
An accuracy of $50\%$ is equivalent to random guessing and represents the lower point of reference.
An accuracy of  $100\%$ corresponds to perfect classification and gives the upper point of reference.

\section{Results}
\label{sec:results}

The performance of the network described in the previous section is evaluated as a function of the prediction
time $T$, the number of training samples $N_T$ and the network size.

\begin{figure}
	\centering
	\includegraphics[scale=1]{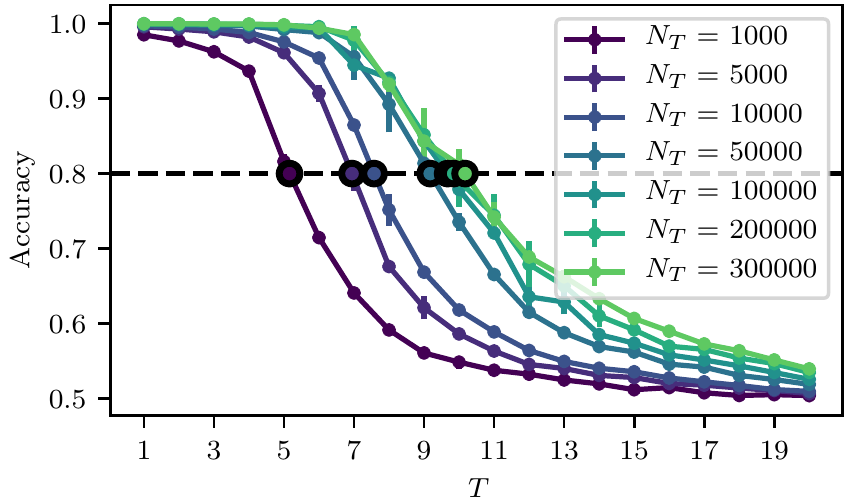}
	\caption{Mean prediction performance. The prediction accuracy is plotted against the prediction horizon $T$ for 
	multiple training datasets. 
	To account for the coverage of the whole chaotic attractor and the stochasticity in the training process, the network is trained three times with different arbitrarily chosen initial seeds and different training sets of size $N_T$ based on the same initial seeds. The resulting variance in the performances is shown as error bars. The dashed black line marks $80\%$ accuracy and the circles with black border are drawn at the horizons for which the model performs with $80\%$ accuracy.}
	\label{fig:single_signals}
\end{figure}

For the results shown in Fig.~\ref{fig:single_signals}, we keep the network topology fixed and study the 
performance as a function of prediction time $T$ and the number of training samples. For short times, the
accuracy of the network is close to $100\%$, but then it quickly deteriorates as $T$ increases and
approaches the random performance of $50\%$. The figure also shows that the performance can be
improved with the number of training samples. Increasing their number from $10^3$ to $3\cdot 10^5$ shows 
that the time interval over which almost perfect prediction is possible increases,
and the point in prediction time where the accuracy decays (defined, e.g., as the point in time where 
the accuracy falls below $80\%$), increases. 


\begin{figure}
	\centering
	\includegraphics[scale=1]{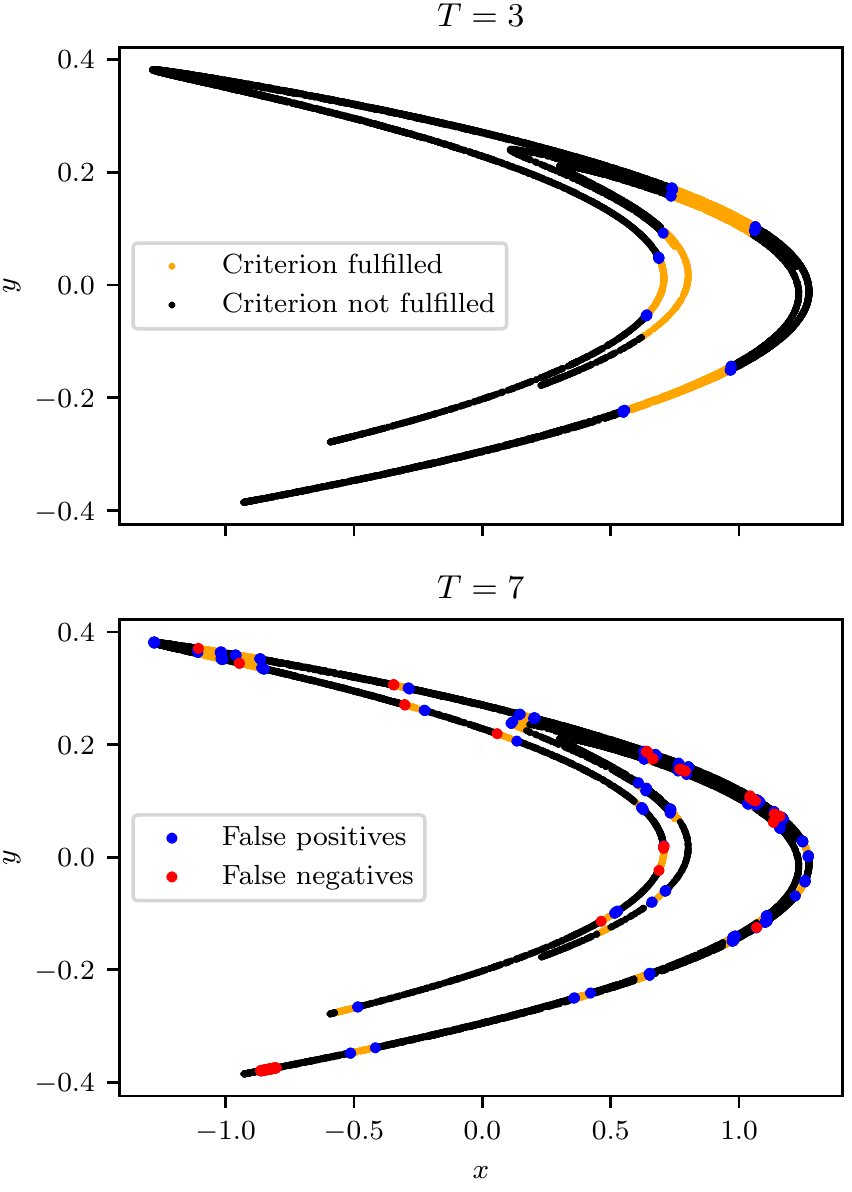}
	\caption{Prediction errors. The leading points $(x_n, y_n)$ that have been misclassified by the neural network for two prediction horizons, $T=4$ and $T=7$. Blue denotes false positives and red denotes false negatives. \review{The orange and black background is the attractor classified into state space points fulfilling and not fulfilling Eq.~\eqref{eq:criterion}, respectively, similar to the color coding in Fig.~\ref{fig:geometrical_task}.}
	}
	\label{fig:prediction_errors}
\end{figure}

Most of the prediction errors occur at the boundaries of the classes, where two classes meet, see Fig.~\ref{fig:prediction_errors}. 
Hence, the expectation is that increasing $T$ leads to less accurate predictions by increasing the number of intervals and therefore the room for errors by insufficient sampling of the regions and their neighborhood. 
Consequently, there is a connection between the prediction horizon and the prediction performance through the number of intervals. 

\begin{figure}
	\centering
	\includegraphics[scale=1]{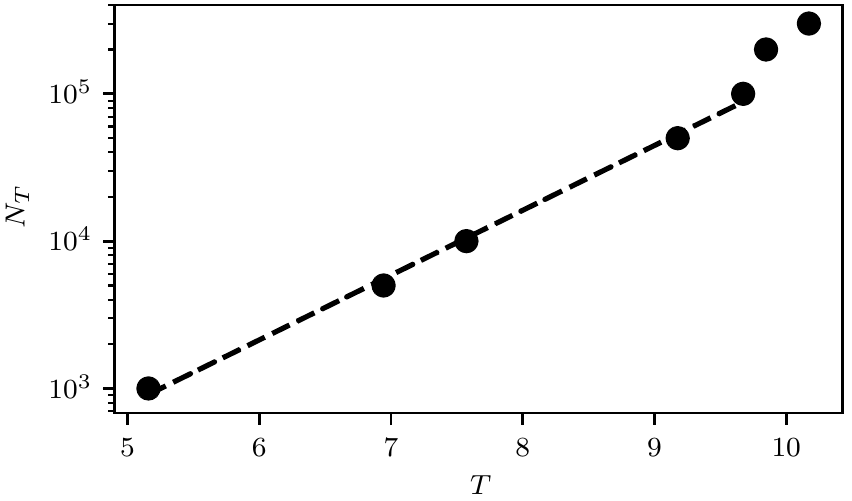}
	\caption{Relation between the number of training samples $N_T$ and the time horizon $T$ 
	up to which the performance is $80\%$ or better. The dashed line shows a fit $N_T\propto \exp(sT)$ with
	$s\approx 1.01$.  For larger times the number of training samples increases more steeply as the 
	capacity of the network has been exceeded. \review{Later in the text, performances of networks with higher capacities are demonstrated.}
	}
	\label{fig:interval_training_samples_relation}
\end{figure}

To analyze the dependence on the number of samples, we take the fixed performance level
of $80\%$, as shown in Fig.~\ref{fig:single_signals}, 
and extract the number of time steps for which accuracy is above this value. 
For instance, for  $N_T=10^3$, this number is $5$, increasing to $7$ with $N_T=5\cdot 10^3$ and $9$ with $N_T=5\cdot 10^4$.
As Fig.~\ref{fig:interval_training_samples_relation} 
shows, the increase in the number of training samples needed to achieve
a certain prediction time increases exponentially, approximately like $40\cdot \exp(1.01 T)$.

Since the H\'{e}non attractor is chaotic, this increase is expected on account of the positive topological entropy $h$.
However, the increase in the number of training samples $s$ is faster than suggested by the topological entropy:
with $h=0.465$ and $s=1.01$, \review{it is around twice as large.}

For very large numbers of training samples the exponential relation breaks down, see Fig.~\ref{fig:interval_training_samples_relation}. For example, using $3\cdot10^5$ training samples does not lead to a six-fold prediction performance improvement over $5\cdot10^4$ training samples. The deviation from the exponential relation in Fig.~\ref{fig:interval_training_samples_relation} indicates that the performance saturates so that more training samples do not lead to higher accuracy. The origin of this saturation lies in the structure and number of
parameters in the network.

\begin{figure}
	\centering
	\includegraphics[scale=1]{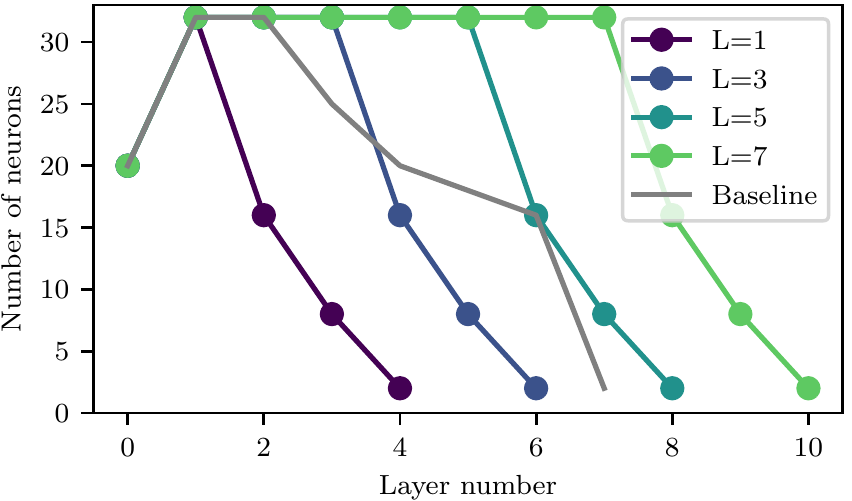}
	\caption{The topologies that are used to check if larger topologies are capable of outperforming smaller ones for larger prediction horizons $T$. 
	}
	\label{fig:topologies}
\end{figure}

To explore the effect of the network, we changed the number of layers and also the number of neurons 
per layer (compared to the previous baseline topology). As there is no ordering on network topologies, we introduce a class of 
networks where the number of input layers is fixed (20, for trajectories with 10 steps into the past), the width
of the first hidden layer is 32, and this is repeated for a certain number of hidden layers, and then there
is an output region of 3 layers, where the number of neurons is (16, 7, 2), with the last one the softmax output
used before.  This decrease from the full width of 32 to the 2 output neurons is faster than in the 
case of the network used before, so as to be able to study smaller networks as well. 
The different networks can then be labeled by $L$, the number of layers with the full width of $32$ neurons,
as shown in Fig.~\ref{fig:topologies}.

\begin{figure}
	\centering
	\includegraphics[scale=1]{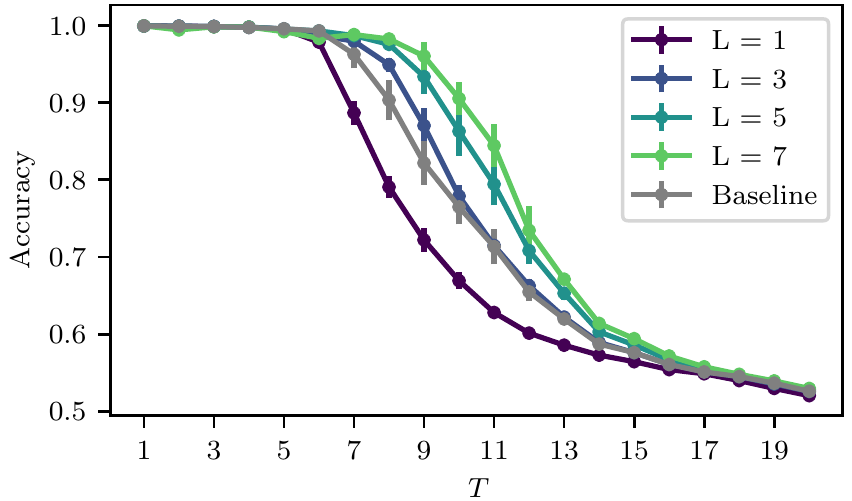}
	\caption{Accuracy of neural networks of different topologies, trained on $10^5$ data samples. 
	The color coding of the topologies corresponds to the one shown in Fig.~\ref{fig:prediction_errors}.
    }
	\label{fig:topologies_results}
\end{figure}

These different network topologies are trained with a fixed number of $10^5$ samples. 
The results are summarized in Fig.~\ref{fig:topologies_results}.
Larger topologies shift the plateau to increased prediction horizons $T$. This can be interpreted as larger networks being able to distinguish finer intervals.

\begin{figure}
	\centering
	\includegraphics[scale=1]{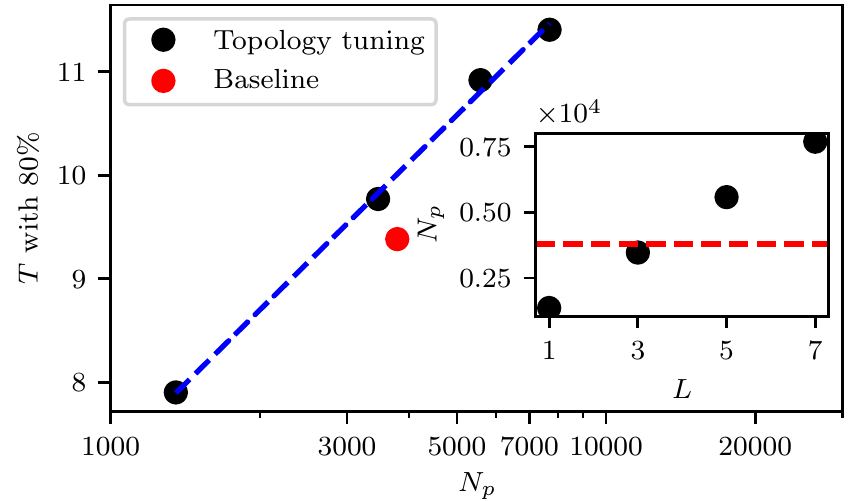}
	\caption{\review{Relation between the time horizon $T$ at which the performance is $80\%$ and the number of parameters in the neural network $N_p$. The relation follows an exponential behavior in $T$. The inset shows the relation between $N_p$ and $L$, the latter as parameter that is used to adjust the number of hidden layers with size $32$. The corresponding properties of the baseline topology are shown in red.}
	}
	\label{fig:topology_tuning_performance}
\end{figure}

Finally, Fig.~\ref{fig:topology_tuning_performance} shows the relation between the number of parameters $N_p$ in the network and the prediction times $T$ at which the different network topologies achieve $80\%$ accuracy. 
The inset shows the linear increase in the number of parameters $N_p$ with the number $L$ of wide hidden layers. 
As the network has to resolve ever finer regions with increasing time, one anticipates that the number of parameters has to increase, and that the increase should be related to the exponentially increasing number of regions.
The data are consistent with a fit $N_p\approx 26 \exp(0.47 T)$, and the exponent $0.47$ is very close to
the topological entropy $h=0.465$. Once the parameters of the network are exhausted, it cannot learn new
features, and no matter how large the training set, the performance will not improve, \review{which is the origin of the deviation from the exponential behaviour in Fig.~\ref{fig:interval_training_samples_relation}
for large times.}

\review{
	Having evaluated the requirements for training data size and neural network size to achieve an accuracy of $80\%$ for $N=10$ state space points as input to the neural network, it is studied in preliminary calculations how the variation of $N$ affects these requirements. Due to the high computational costs, the baseline neural network is selected as network topology and trained a single time with $N\in[1, 15]$ for different prediction times $T$ and training data sizes $N_T$, similar to Fig.~\ref{fig:single_signals}. For every value of $T$ and $N_T$, the performance increases while tuning $N$ from $15$ down to $1$. Hence, all curves in Fig.~\ref{fig:single_signals} shift towards the right for smaller values of $N$. The performance increase, however, appears to depend on $N_T$: the performance differences between the trainings with different values of $N$ become smaller for more training data $N_T$. Consequently, the exponential scaling factor $s$ as computed in Fig.~\ref{fig:single_signals} depends on $N$. Nevertheless, they remain positive for all $N$ so that ever more training data is needed to achieve an accuracy of $80\%$.
	}


\section{Conclusions}
\label{sec:conclusions}

In the present work, we have analyzed the performance of neural networks for predictions
in a chaotic 2d model. The benefit of the 2d map is the ability to visualize the state space of the
system and to given a geometric interpretation to the prediction task: the points on the
attractor that fall above the prediction threshold are part of geometrically well defined
rectangular regions in the state space. As the prediction time increases, so does their number,
while their diameter decreases. The network therefore has to classify initial conditions according
to whether they fall into these regions or not.

As was to be expected, the performance of sufficiently complex network topologies  
was very good for a few steps into the future 
and then deteriorated quickly as the prediction time
increased. The prediction errors are localized near the boundaries of the regions, and reflect the fact
that a finite number of training samples can only approximate the regions and their boundaries within
a certain accuracy. Accordingly, the performance could be improved with an increase in the number 
of training samples. The number of training samples needed to maintain a certain performance increased more quickly
than the number of regions that have to be \review{distinguished}. 

Increasing the number of training samples increases the performance only up to a point. 
Eventually, another limitation of the prediction task becomes noticeable, the fact that a finite number
of parameters in the network cannot characterize a much larger number of regions, unless
they are in some form correlated. Then an improvement in performance can only be achieved
by an increase in the number of nodes in the network.

While accuracy and number of training samples are parameters that can easily be ordered
as to good and bad or large and small, the network topology remains a variable without
natural ordering, and with an irregular influence on the performance. The family of
networks with an increasing number of hidden layers of fixed size did show an improved performance
with an increase in the number of hidden layers. But it remains to be studied how variations
in the width or  the shape of the tapering region towards the output region influence the 
performance. The results in Fig.~\ref{fig:topologies_results} show that dependence is non-monotonic:
the baseline model has a tapering region with five layers, and performance similar to the $L=3$ model
with a steep decent towards the output layer. Nevertheless, within a fixed topology, one can find
a direct relation between the number of parameters in the network and the number of regions that have
to be identified, and both increase exponentially with a rate given by the topological entropy.

In addition to the network topology and the training protocol, also the time history of the
trajectory is an important parameter. 
The results shown here are for a history of $N=10$ steps into the past.
\review{
	First tests for values of
$N$ between $1$ and $15$ show that smaller values improve the prediction performances but leaving the
main result of exponentially larger training data corpus sizes for longer prediction time horizons in the H\'{e}non
map unchanged. The improvement for smaller $N$ are reasonable when taking into account that the
Henon map is fully observed and a deterministic chaotic map.
However, it should be kept in mind that when $N$ is changed, also the input layer of the network topology
changes while the rest of the network remains unchanged in our setup. This results in more network
parameters $N_p$ per input feature and, hence, an increased network capacity per input feature for
smaller numbers of $N$.
}

The present study is performed for a particular system and for a well-defined numerical task. Nevertheless,
the results should also carry over to more general, non-numerical ML tasks, where the number of 
regions correspond to certain regions in the state space of the system: decisions that depend on 
regions that are under-resolved or not identified at all by the training sets will naturally
be error prone.

The results presented here in conjunction with observations in \cite{pathak2018model} show that Machine Learning ideas
provide a set of flexible and easily adapted tools for predicting chaotic dynamical systems, if 
basic elements of the underlying chaotic dynamics are taken into account, such as the increase
in complexity of the task with the prediction time, and the fragmentation of state space by
the stretch and fold mechanism. When these conditions are taken into account, 
ML tools hold great promise for many applications.


\review{
One such application in the context of chaotic dynamics is the predicition
of relaminarization events in wall-bounded parallel shear flows such as pipe,
Couette or channel flow.  The transition to turbulence in such flows is
subcritical, resulting in the coexistence of an attracting fixed point
corresponding to laminar flow and a chaotic saddle representing transient
turbulent dynamics.  A relaminarization event then corresponds to an escape
from the chaotic saddle.  A trajectory will therefore never return to the
chaotic saddle once it has reached the laminar fixed point, unless it is forced
to do so by a finite amplitude perturbation. In the present context this
implies that once a relaminarization event has been recorded, a new flow
simulation will need to be initialised within the chaotic saddle in order to
record the next event. That is, the training process requires a large number of
simulations.}

\review{
The present study confirms the general feasibility of data-driven
predictions of extreme events despite chaotic dynamics.  However, the input spaces in the fluid dynamics
context are of much higher dimension and as such require higher capacities of
the neural networks to achieve good prediction results. Additionally, the
prediction horizon will have to be chosen with care to ensure sufficiently high
performances at non-trivial task difficulty. After all, high neural network
performances are required to, eventually, learn previously unknown facts from
the neural network.
}


\section{Acknowledgements}

We gratefully thank Bruno Eckhardt for his supervision for this project and all the personal and professional advice along our ways, from studies and far beyond. All this will be remembered.



\bibliography{henon_prediction}

\end{document}